\documentclass[letterpaper, 10 pt, conference]{ieeeconf}  

\IEEEoverridecommandlockouts                              
\usepackage{graphicx}
\usepackage{amsmath}
\usepackage{algorithm}
\usepackage{algpseudocode}
\usepackage{booktabs}
\usepackage{multirow}
\usepackage{amssymb}
\usepackage{adjustbox}
\usepackage{xcolor}
\usepackage{url}
\usepackage{hyperref}
\hypersetup{
    colorlinks=true,
    urlcolor=blue,
    linkcolor=blue,
    citecolor=blue
}
\title{\LARGE \bf
SUREFlow: State-space Uncertainty-aware REsidual Flow Matching for Robust Robot Manipulation}
\author{Md Tanvir Islam, Sai Navaneet Peddapalli, Sangmoon Lee, Sangtae Ahn* 
\thanks{Md Tanvir Islam, Sai Navaneet Peddapalli, Sangmoon Lee, and Sangtae Ahn are with School of Electronic and Electrical Engineering, Kyungpook National University, Daegu 41566, Republic of Korea. (e-mails: tanvirnwu, navaneet, moony, stahn @knu.ac.kr)}
\thanks{*Corresponding author: Sangtae Ahn (stahn@knu.ac.kr)}
}

\begin{document}

\maketitle

\thispagestyle{empty}
\pagestyle{empty}

\begin{abstract}
Generative vision-language-action policies have advanced robot manipulation, but they often exhibit instability under noise, partial observability, and stochastic initial conditions. During extended rollouts, small velocity errors accumulate, degrading execution reliability. Existing diffusion and flow-based policies typically assume homoscedastic residuals and lack explicit uncertainty modeling within action generation, limiting robustness during iterative rollout.
We propose SUREFlow, a state-space uncertainty-aware residual flow matching framework built on a Mamba backbone. The method jointly predicts action velocities and input-dependent residual uncertainty, enabling selective refinement of unreliable action dimensions without environment feedback while preserving computational efficiency.
On LIBERO, SUREFlow achieves 92.5\% average success rate (SR), outperforming the Mamba-based MaIL by 34.2\%. On LIBERO-PRO, it attains around 49\% SR using only 179M parameters, achieving performance comparable to large VLAs with 3-7B parameters. SUREFlow source code is available on: \href{https://github.com/tanvirnwu/SUREFlow}{https://github.com/tanvirnwu/SUREFlow}.

\end{abstract}

\vspace{0.25cm}
\section{Introduction}
Learning robot manipulation (RM) policies from demonstrations has become a dominant paradigm, where recent vision-language-action (VLA) models leverage multi-view visual observations and natural language conditioning to enable generalizable robotic skills across diverse tasks~\cite{cao2025mamba, zitkovich2023rt, jia2024mail}.
Despite this progress, many RM approaches still rely on direct action regression formulations~\cite{florence2022implicit}, which yield deterministic predictions and are known to struggle when demonstrations exhibit multimodal behaviors, as commonly observed in real-world tasks~\cite{lynch2020learning}.

To address these limitations, generative policy learning methods have recently gained attention. Diffusion-based and flow-based policies model the conditional distribution over actions rather than predicting a single output, enabling stochastic action generation and improved expressiveness, particularly in robotic manipulation settings \cite{cao2025mamba, chi2025diffusion, lipman2023flow}. Such methods have demonstrated improved robustness and flexibility compared to deterministic baselines. However, purely continuous generative policies typically operate at a single temporal scale and lack explicit temporal abstraction. As a result, stochasticity introduced during generation may accumulate over long action sequences, leading to unstable execution and inconsistent behaviors~\cite{lynch2020learning,chi2025diffusion}. Moreover, these approaches typically incur high computational costs due to iterative sampling and often rely on transformer-based architectures, which substantially increase computational costs.
\begin{figure}[t]
    \centering
    \includegraphics[width=0.48\textwidth]{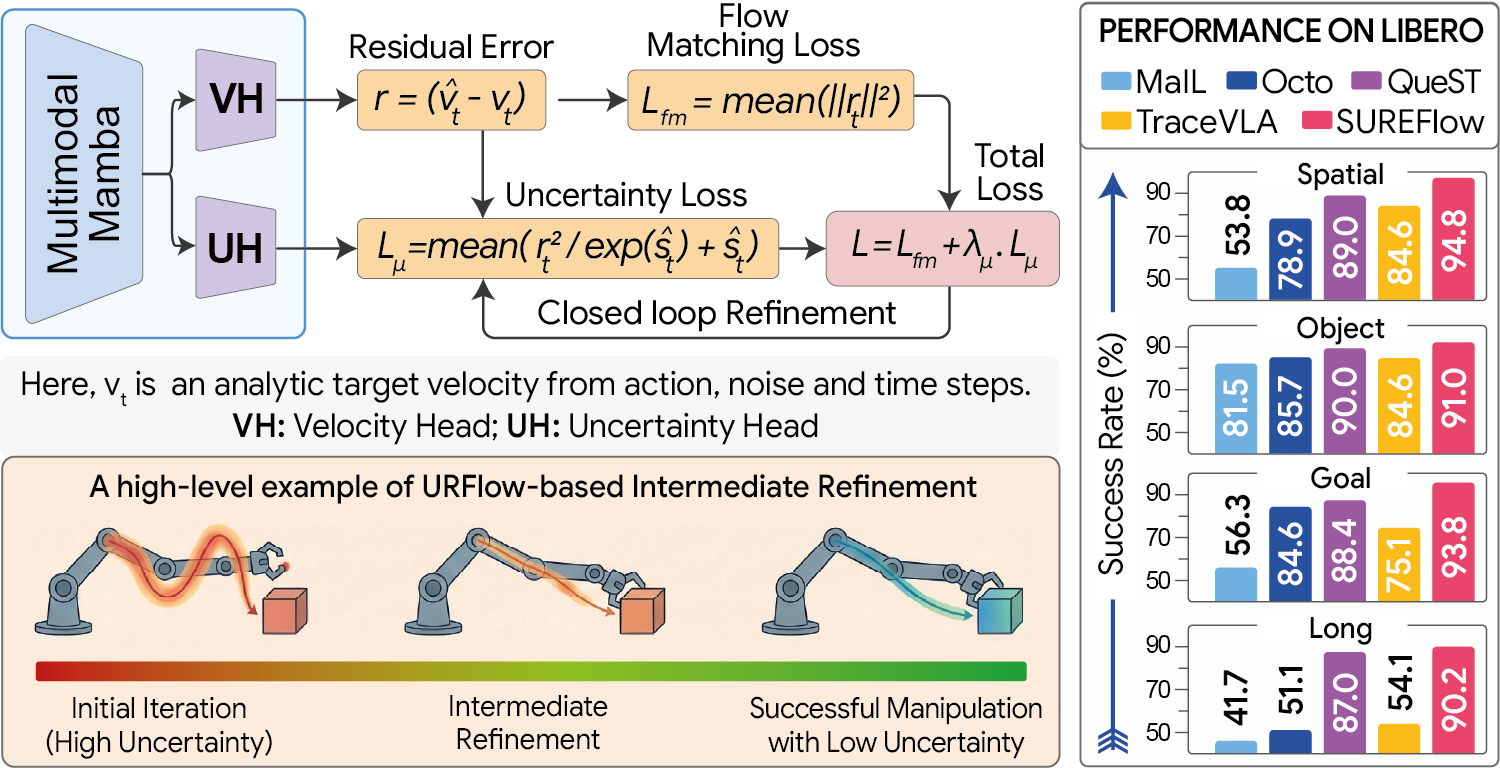}
    \caption{Overview of URFlow, illustrating closed-loop refinement of uncertain action dimensions via internal residual updates during inference, without external feedback. The right panel shows representative LIBERO results, demonstrating consistent improvements over recent SOTA baselines.}
    \label{fig:fig11}
\end{figure}

In parallel, prior work has explored hierarchical and skill-based representations for RM, aiming to learn reusable skills directly from data rather than relying on hand-crafted primitives~\cite{ma2024hierarchical, liutextit}. However, most such approaches either separate skill learning from low-level control or depend on explicit planning modules, limiting seamless integration with end-to-end generative policies.
Meanwhile, alternative sequence modeling architectures have recently gained attention for long-horizon reasoning. State-space models such as Mamba~\cite{gu2024mamba} offer linear time complexity and strong temporal modeling capabilities, providing an efficient alternative to large backbones like Transformers. Despite their success in language and vision domains, their application to generative robot control remains limited. Moreover, generative manipulation policies introduce stochasticity during action generation, which can accumulate over long-horizon rollouts and lead to unstable execution when actions are applied open-loop. Existing approaches typically lack explicit mechanisms to quantify or correct uncertain action predictions during inference, limiting their robustness in complex manipulation tasks utilizing a comparatively lightweight architecture.

In this work, we address this gap by proposing \textit{``State-space Uncertainty-aware REsidual Flow (SUREFlow)"}, a manipulation policy based on Mamba~\cite{gu2024mamba} that tackles long-horizon instability through action flow matching, efficient sequence modeling, and uncertainty-aware inference. The proposed framework formulates action generation as a continuous-time flow matching problem and learns a velocity field that maps noise to actions conditioned on visual observations, robot states, and task embeddings. To improve robustness during execution, it incorporates uncertainty-aware modeling that estimates prediction confidence alongside the action velocity field, as illustrated in Fig.~\ref{fig:fig11}. During inference, predicted uncertainty is used to selectively refine action sequences, mitigating error accumulation and enable stable long-horizon rollout execution, while maintaining a lightweight multimodal backbone and consistently outperforming recent state-of-the-art (SOTA) robot manipulation baselines across multiple benchmarks.
Our key contributions are as follows:

\begin{itemize}

    \item We introduce an \textit{Uncertainty-aware Residual Flow (URFlow)} refinement framework within conditional flow matching for generative robot manipulation. URFlow learns input-dependent velocity variance and selectively corrects unreliable action dimensions with negligible computational overhead, stabilizing iterative flow integration under extended rollouts.
    
    \item We propose integrating a lightweight \textit{Memory-Guided Action Decoder (MGAD)} that re-attends learnable action queries to multimodal memory representations, enabling context-aware and structured action refinement. This mechanism enhances temporal conditioning and improves robustness in generative action synthesis, as validated through ablation studies.

    \item To the best of our knowledge, integrating uncertainty-aware residual refinement within a state-space generative flow policy has not been previously explored. We address this gap by developing \textit{SUREFlow}, a Mamba-based VLA architecture that unifies conditional flow matching with URFlow and MGAD. Leveraging linear-time sequence modeling, SUREFlow enables scalable and efficient generative control with only 179M parameters, avoiding reliance on large multimodal backbones.

\end{itemize}

\begin{figure*}[t]
    \centering
    \includegraphics[width=1.0\textwidth]{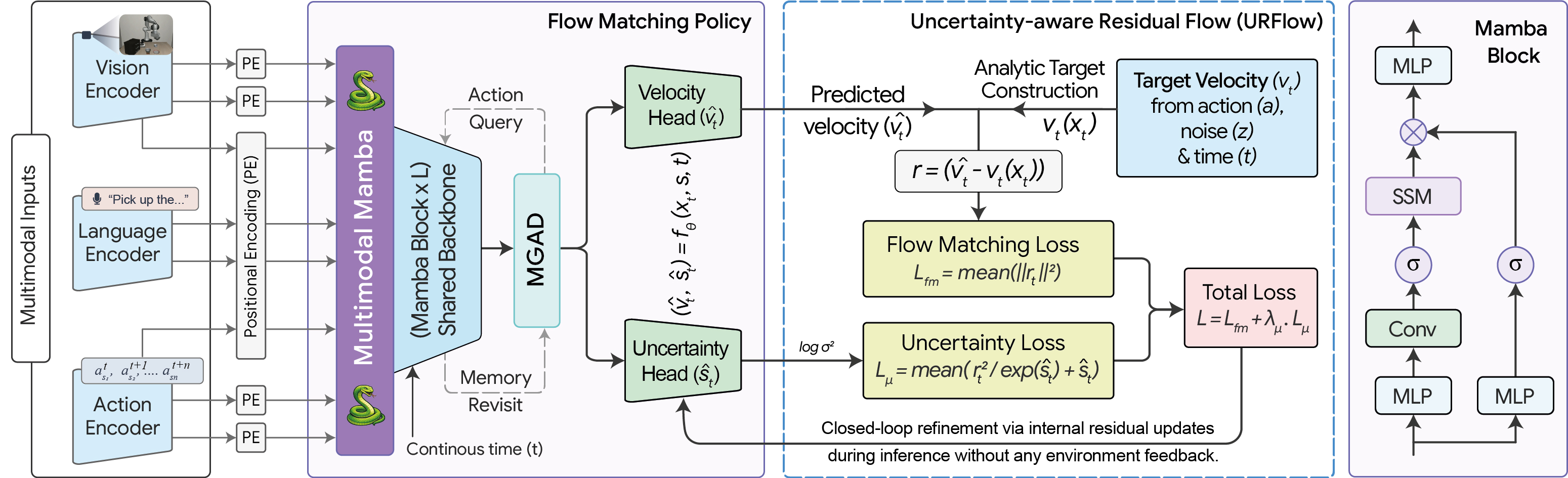}
   \caption{Overall architecture of our proposed \textit{SUREFlow}. Multimodal inputs are encoded and fused into a unified token representation. The fused tokens are processed by a shared backbone (stack of L Mamba blocks) conditioned on continuous flow time with the \textit{`Memory-Guided Action Decoder (MGAD)'} module to enhance temporal conditioning and contextual reasoning. The flow-matching policy predicts an action velocity field and an uncertainty estimate, yielding our Uncertainty-aware Residual Flow (URFlow), which constructs flow-matching and uncertainty-weighted losses during training. At inference time, URFlow performs closed-loop refinement via internal residual updates guided by predicted uncertainty, improving robustness during long-horizon execution without online action correction or environment feedback.}
    \label{fig:SUREFlow}
\end{figure*}

\section{Related Work}
\subsection{VLAs for Robot Manipulation}
Recent advances in VLA models have significantly expanded the scope of generalist robotic manipulation by leveraging large-scale vision-language pretraining. Early foundation models such as RT-1~\cite{brohan2022rt} and RT-2~\cite{zitkovich2023rt} demonstrated that pretrained vision-language representations can be adapted for discretized robot action prediction, enabling zero-shot generalization across diverse tasks and embodiments. Building on this paradigm, OpenVLA scaled to 7B parameters~\cite{kim2024openvla}, while Octo trained a transformer-based generalist policy across heterogeneous trajectories to promote multi-embodiment generalization~\cite{octo_2023}.
Subsequent efforts focused on efficiency and architectural refinement. TinyVLA reduced model size while retaining competitive performance~\cite{wen2024tinyvla}, and $\pi_{0.5}$~\cite{black2025pi} improved deployment efficiency of pretrained VLA backbones. QueST introduced structured tokenization and query-based conditioning~\cite{mete2024quest}, whereas TraceVLA injected spatial-temporal visual traces without modifying the backbone~\cite{zheng2025tracevla}. In parallel, Mamba~\cite{gu2024mamba} has been explored as an efficient alternative to transformer backbones. MaIL~\cite{jia2024mail} leverages Mamba for long-horizon imitation learning, and RoboMamba~\cite{liu2024robomamba} integrates Mamba within a VLA framework for multimodal policy learning.

Beyond autoregressive VLAs, generative policies have emerged as an alternative paradigm for action synthesis. Diffusion-based approaches such as Diffusion Policy~\cite{chi2025diffusion} and 3D Diffusion Policy~\cite{Ze2024DP3} model multimodal action distributions through iterative denoising, but require multiple sampling steps during inference. Flow-based methods, including Flow Matching~\cite{lipman2023flow} and continuous normalizing flows~\cite{albergo2023building}, instead learn velocity fields that transport a prior distribution toward expert actions without explicit diffusion. While promising for generative modeling, these formulations typically assume homoscedastic residuals and lack explicit uncertainty-aware refinement mechanisms for stabilizing extended-horizon rollout execution.

Despite these advances, many VLA models rely on large transformer backbones and autoregressive decoding, resulting in substantial computational overhead. Furthermore, explicit modeling of input-dependent uncertainty is typically absent, reducing robustness during iterative rollout and extended rollout execution. These limitations motivate exploring efficient state-space backbones~\cite{gu2024mamba} combined with uncertainty-aware generative refinement to improve scalability and extended-horizon execution stability.

\subsection{Uncertainty Modeling and Residual Learning}
Uncertainty estimation has been widely studied in deep learning and control. Kendall and Gal~\cite{kendall2017uncertainties} introduced heteroscedastic regression for modeling input-dependent variance, while uncertainty-aware reinforcement learning (RL) improves exploration and safety~\cite{liu2022uncertainty, depeweg2018decomposition}. However, these approaches are generally applied to value estimation or supervised prediction rather than generative action transport. Additionally, residual learning has proven effective in stabilizing robotic control. Residual RL~\cite{johannink2019residual} combines model-based controllers with learned residual policies to correct systematic errors. Related strategies learn additive corrections to mitigate model mismatch, yet such residual formulations have not been integrated with flow-based generative policies.

\section{Problem Formulation}
\label{sec:problem_formulation}
We consider the task of learning a goal-conditioned robot manipulation policy $\pi(a \mid o, g)$ that maps multimodal observations to a sequence of continuous actions. Here, $o \in \mathcal{O}$ denotes the current observation, $g$ is a natural 
language task instruction, and $a \in \mathcal{A}$ is the target action sequence. The observation consists of multi-view RGB images and robot proprioceptive states as follows:
\begin{equation}
\footnotesize
o_t = \{ I_t^{(1)}, I_t^{(2)}, s_t \},
\vspace{-0.2cm}
\end{equation} 
where $I_t^{(1)}$ and $I_t^{(2)}$ denote the agent's view and eye-in-hand RGB images, respectively, and $s_t$ denotes the robot joint and gripper states. The task instruction $g$ is encoded as a fixed task embedding $e$ using a precomputed language encoder.

In our proposed SUREFlow, we formulate policy learning through the lens of conditional flow matching. Rather than directly regressing actions, the objective is to learn a time-dependent vector field $v_t$ that induces a 
probability path $p_t(x)$ transforming a simple prior distribution $p_1$ 
(Gaussian noise) into the expert action distribution $p_0$.
To construct this path, we define a conditional probability trajectory $x_t$ for a continuous time variable $t \in [0,1]$ as a linear interpolation between a ground truth action $a \sim p_0$ and a noise sample $z \sim \mathcal{N}(0, I)$ as follows:
\begin{equation}
\footnotesize
\label{eq:probability_path}
x_t = (1 - t)a + tz .
\vspace{-0.2cm}
\end{equation}

The analytic vector field that generates this transformation is given by the target velocity as $v_{\text{t}}(x_t) = z - a$. We model the learned conditional velocity field as $\hat{v}_t = f_\theta(x_t, s, t)$, where $s = \psi(o_t, e)$ denotes the fused multimodal conditioning representation.
Unlike standard formulations, \textit{SUREFlow} treats the discrepancy between the predicted and analytic target velocity as a residual flow matching problem. We hypothesize that this residual captures informative structure related to task complexity and data uncertainty, which can be exploited to refine policy behavior.

\section{Proposed Method: SUREFlow}
\label{sec:method}
We present SUREFlow, an end-to-end VLA framework that integrates uncertainty-aware residual flow matching within a unified state-space backbone. 
The overall architecture is illustrated in Fig.~\ref{fig:SUREFlow}. Given multimodal inputs comprising RGB observations, robot proprioceptive states, and task language embeddings, each modality is encoded into a shared latent space through dedicated encoders and multimodal fusion. The resulting token sequence is processed by a Mamba-based state-space backbone, enabling efficient modeling of long-range temporal dependencies.

On top of this backbone, a flow matching head predicts a conditional velocity field that transports Gaussian noise toward expert action trajectories. In parallel, an auxiliary head estimates input-dependent uncertainty over the predicted residual. Unlike conventional diffusion or flow-based policies that optimize uniform velocity residuals, SUREFlow explicitly models heteroscedastic uncertainty, formulating action generation as a residual learning problem. This design supports training-time closed-loop refinement and enables uncertainty-guided closed-loop residual refinement during inference, minimizing the discrepancy between the predicted and analytic target velocities.
To further enhance conditioning and expressiveness, the framework optionally incorporates Flow Conditioned Modulation (FCM) for timestep-dependent feature modulation and MGAD to ground learnable action queries in multimodal observation memory.

\subsection{Encoding and Conditioning}
The first stage of SUREFlow is responsible for transforming raw multimodal inputs into a unified latent representation suitable for conditional flow matching. At each time step $t$, the observation $o_t$ contains two RGB images $I_t^{(1)}$ and $I_t^{(2)}$ corresponding to the agent view and the eye-in-hand view. 
Each image is processed independently using a convolutional visual encoder. Formally, for each camera view $k \in \{1,2\}$, we compute the visual embedding as follows:
\begin{equation}
\footnotesize
z_t^{(k)} = f_{\text{img}}^{(k)}(I_t^{(k)}) \in \mathbb{R}^{d},
\vspace{-0.1cm}
\end{equation}
where $f_{\text{img}}^{(k)}$ denotes a ResNet-18 backbone with Group Normalization, and $z_t^{(k)}$ is the resulting visual embedding. This produces one visual token per camera view at each $t$.

The robot joint and gripper states $s_t$ are embedded using a linear projection to obtain a proprioceptive token defined as $z_t^{(s)} = f_{\text{state}}(s_t)$, where $f_{\text{state}}$ is a learnable affine mapping. This token captures the instantaneous physical configuration of the robot and complements the visual observations. The natural language instruction $g$ is encoded using a pretrained language encoder to obtain a fixed task embedding $e$, which remains constant throughout the trajectory and serves as a global conditioning signal for the manipulation goal. To incorporate the continuous flow time variable $t \in [0,1]$ introduced in Sec.~\ref{sec:problem_formulation}, we use a learnable time embedding $\phi(t) \in \mathbb{R}^{d}$. This embedding enables the policy to adapt its behavior along the probability path connecting the noise distribution $p_1$ and the expert action distribution $p_0$. The encoded components are combined to form a unified token sequence at time step $t$ as follows:
\begin{equation}
\footnotesize
\label{eq:conditioned_token}
X_t = \left[ \phi(t),\; e,\; z^{(1)}_t,\; z^{(2)}_t,\; z^{(s)}_t,\; x_t \right],
\vspace{-.1cm}
\end{equation}
where $x_t$ denotes the current action token along the probability path defined in Sec.~\ref{sec:problem_formulation}. This sequence aggregates all conditioning information required for action generation, including task context, perception, robot state, and flow time. This unified token sequence $X_t$ serves as the input to the sequence modeling component described in the following subsection, where temporal dependencies and conditional action generation are performed.

\subsubsection{Flow Conditioned Modulation (FCM)}

Beyond explicit concatenation of the flow-time token, SUREFlow employs FCM to inject temporal conditioning directly into intermediate feature representations. FCM leverages the time embedding $\phi(t)$ to perform feature-wise affine modulation, enabling smooth conditioning of the sequence model without modifying the token structure. In practice, $\phi(t)$ is implemented using a sinusoidal frequency embedding followed by a two-layer MLP, producing a $d$-dimensional vector.

Let $U \in \mathbb{R}^{B \times N \times d}$ denote a sequence of $N$ token features with embedding dimension $d$, where $U$ may represent state tokens, action tokens, task tokens, or decoded action features. From the flow-time embedding $\phi(t)$, we extract a conditioning vector $c \in \mathbb{R}^{B \times d}$ (selecting the first token when represented as a sequence). Two learnable linear projections generate feature-wise scale and shift parameters as follows:
\begin{equation}
\footnotesize
\gamma = W_{\gamma} c + b_{\gamma}, 
\qquad 
\beta = W_{\beta} c + b_{\beta},
\vspace{-0.2cm}
\end{equation}
where $\gamma, \beta \in \mathbb{R}^{B \times d}$. The modulated representation is computed via broadcasting across the token dimension:
\begin{equation}
\small
\mathrm{FCM}(U, \phi(t)) = U \odot (1 + \gamma) + \beta,
\vspace{-.15cm}
\end{equation}
where $\odot$ denotes element-wise multiplication and $\gamma, \beta$ are broadcast along the token dimension.
To ensure stable optimization, all modulation parameters are initialized to zero, yielding $\gamma = 0$ and $\beta = 0$ at initialization, such that FCM initially acts as an identity mapping. 

In SUREFlow, FCM is applied to embedded state tokens, embedded action tokens, and decoded action features prior to the final prediction head. When goal conditioning is enabled, it is also applied to the task embedding. This design introduces continuous flow-time awareness throughout the network while preserving architectural simplicity.

\subsection{Sequence Modeling and Action Generation}
\label{sec:seq_model}
Given the conditioned token sequence $X_t$ constructed in Eq.~(\ref{eq:conditioned_token}), \textit{SUREFlow} performs joint sequence modeling and conditional action generation using a shared Mamba~\cite{gu2024mamba} backbone. This component is responsible for integrating multimodal context, temporal structure, and flow time conditioning to predict the action velocity field required for conditional flow matching.

The token sequence $X_t$ is processed by a stack of $L$ layers, which serve as the shared sequence backbone of the policy. Each layer implements a selective State-space model (SSM)~\cite{gu2024mamba} that updates hidden states through linear recurrent dynamics combined with input-dependent gating. SSM enables efficient modeling of long-range temporal dependencies with linear computational complexity, making it well-suited for long-horizon execution and temporal modeling. If we formally let $H_t^{(0)} = X_t$ to denote the input token sequence, the sequence backbone is computed as follows:
\begin{equation}
\small
H_t^{(\ell)} = \mathrm{Mamba}_{\ell}\big(H_t^{(\ell-1)}\big), \quad \ell = 1, \dots, L,
\end{equation}
where $H_t^{(\ell)}$ denotes the hidden token representations at layer $\ell$. The final representation $H_t^{(L)}$ encodes joint information from visual observations, robot states, task context, flow time, and the current action token.

\subsubsection{Memory-Guided Action Decoder (MGAD)}
To further enhance the interaction between action representations and multimodal context, SUREFlow optionally employs MGAD that treats the action token as a learnable query and allows it to attend to the remaining tokens in $H_t^{(L)}$, including visual, proprioceptive, and task tokens. This mechanism enables the action representation to selectively aggregate relevant contextual information before velocity prediction.

Let $h_t^{(a)}$ denote the action token extracted from $H_t^{(L)}$. The MGAD module in SUREFlow updates this token as follows:
\begin{equation}
\small
\tilde{h}_t^{(a)} = \mathrm{MGAD}\big(h_t^{(a)}, H_t^{(L)}\big),
\end{equation}
where $\mathrm{MGAD}(\cdot)$ denotes a cross-attention module parameterized by the policy. 
Unlike full transformer decoders that employ multiple learnable action queries and stacked decoder layers, MGAD performs a lightweight, single-step cross-attention refinement over the action token, preserving architectural efficiency. When MGAD is not used, the original action token $h_t^{(a)}$ is passed directly to both the \textit{policy head} and the \textit{uncertainty head}.

\subsubsection{Velocity and Uncertainty Prediction}

The refined action representation is mapped to the parameters required for conditional flow matching. Specifically, from the decoded action features $\tilde{h}_t^{(a)}$, the policy predicts both the velocity and the associated uncertainty as follows:
\begin{equation}
\small
\label{eq:predictions}
\hat{v}_t = f_v\!\left(\tilde{h}_t^{(a)}\right), 
\qquad
\hat{s}_t = f_s\!\left(\tilde{h}_t^{(a)}\right),
\end{equation}
where $f_v$ and $f_s$ are linear projection heads with independent parameters. 
Together with the shared backbone defined in Eq.~(7), they instantiate the overall policy 
$f_\theta(x_t, s, t)$, where the backbone produces the decoded action representation 
$\tilde{h}^{(a)}_t$ and the projection heads map it to velocity and uncertainty predictions. 
The predicted velocity $\hat{v}_t$ therefore corresponds to the learned conditional vector 
field $f_\theta(x_t, s, t)$, while $\hat{s}_t$ estimates the log-variance 
($\log \sigma^2$) of the residual and is used for uncertainty-aware modeling and refinement.

\subsection{Uncertainty-aware Residual Flow (URFlow)}
\label{sec:URFlow}
Standard conditional flow matching learns a deterministic approximation of the analytic target velocity, implicitly treating all residual errors as equally reliable. During extended rollout execution, demonstrations can be noisy or ambiguous, and the reliability of the learned flow may vary across action dimensions and contexts. SUREFlow addresses this limitation by introducing an uncertainty-aware residual formulation within conditional flow matching, inspired by heteroscedastic regression for modeling input-dependent uncertainty~\cite{kendall2017uncertainties} and residual learning strategies for stabilizing robot control~\cite{johannink2019residual}. URFlow explicitly models heteroscedastic uncertainty in the learned flow velocity field, enabling selective refinement of unreliable action dimensions and mitigating compounding integration errors during iterative flow execution.

\subsubsection{Residual Flow Definition}
As defined in Sec.~\ref{sec:problem_formulation}, the probability path is $x_t$ with $a \sim p_0$ and $z \sim \mathcal{N}(0,I)$, and the analytic target velocity is $v_{\text{t}}(x_t)$. Specifically, our policy $f_{\theta}$ predicts both the velocity $\hat{v}_t$ and its uncertainty $\hat{s}_t$ via a function that is shown in line 4 of Alg.~1. This is instantiated via a shared backbone defined in Eq.~(7), followed by the projection heads $f_v$ and $f_s$ defined in Eq.~(9).  Given the predicted velocity $\hat{v}_t$ produced by the policy $f_{\theta}$, we define the residual flow as $r_t$, as in line 5 of Alg.~1.

\begin{algorithm}[!t]
\footnotesize \label{alg}
\caption{Pseudocode of URFlow in our SUREFlow.}
\label{alg:URFlow}
\begin{algorithmic}[1]
\Require Action $a \sim p_0$, noise sample $z$, flow time $t$, observation $o_t$, instruction $g$, task embedding $e$, policy $f_{\theta}$, uncertainty weight $\lambda_u$, threshold $\tau$, refinement steps $K$, step size $\eta$.
\Ensure Training loss $\mathcal{L}$ and refined action sequence $a$.

\State Encoding: compute fused latent $s = \psi(o_t, e)$
\State Path construction: $x_t \gets (1-t)a + tz$
\State Target velocity: $v_{\text{t}}(x_t) \gets z - a$

\State Predictions: $(\hat{v}_t, \hat{s}_t) \leftarrow f_{\theta}(x_t, s, t)$
\State Residual: $r_t \gets \hat{v}_t - v_{\text{t}}(x_t)$
\State Flow matching loss: $\mathbb{E}_{a,z,t}\left[\left\| r_t \right\|_2^2\right]$
\State Uncertainty loss: $\mathcal{L}_{u} \gets \|r_t\|_2^2 / \exp(\hat{s}_t) + \hat{s}_t$
\State Total loss: $\mathcal{L} \gets \mathcal{L}_{\mathrm{FM}} + \lambda_u \mathcal{L}_{u}$

\State return $\mathcal{L}$ \Comment{\textit{Used during training}}

\vspace{0.8em}
\Statex \textbf{Inference time uncertainty-aware action refinement}

\For{$k = 1$ to $K$}
    \State At $t=0$: $(\hat{v}_0,\hat{s}_0) \gets f_{\theta}(a, s, 0)$ 
\State $\mathcal{M} \gets \{ i \mid \hat{s}_{0,i} > \tau \}$ \Comment{\textit{Compute uncertainty mask}}
    \ForAll{$i \in \mathcal{M}$}
        \State $a_i \gets a_i - \eta \hat{v}_{0,i}$
    \EndFor
\EndFor
\State return refined $a$
\end{algorithmic}
\end{algorithm}

\subsubsection{Training Objective}
A standard flow matching loss is then optimized together with an uncertainty-aware residual loss $\mathcal{L}_{\mathrm{u}}$ as in Alg.~1 (line 7).
The uncertainty-aware residual loss models $r_t$ as heteroscedastic and penalizes residuals under the predicted variance. The final training objective is defined in Alg.~1 (line~8), where $\lambda_u$ controls the contribution of the uncertainty term.

\subsubsection{Inference Time Refinement}
At inference time, the policy generates an action sequence by integrating the learned conditional velocity field $\hat{v}_t$ predicted by the policy to transport an initial noise sample toward the action distribution. 
To enhance stability in extended rollout steps, URFlow performs a deterministic refinement procedure guided by the predicted uncertainty $\hat{s}_t$ introduced in Eq.~(\ref{eq:predictions}). Given a generated action sequence $a$, we evaluate the predicted uncertainty at $t=0$ and construct a mask over unreliable action dimensions as defined in line 12 of Alg.~1,
where $\tau$ is a predefined uncertainty threshold and $\hat{s}_{0,i}$ denotes the $i$-th dimension of the predicted log-variance at $t=0$. 
For dimensions $i \in \mathcal{M}$, the action is refined using the predicted velocity at $t = 0$ via $a_i \leftarrow a_i - \eta \hat{v}_{0,i}$, where $\eta$ is a fixed step size.
This update is applied for a small number of refinement steps, and it operates entirely within the model without any online environment feedback or action correction.

\begin{table*}[t]
\centering  \scriptsize
\renewcommand{\arraystretch}{1.0}
\caption{Performance comparison on LIBERO and Meta-World benchmarks. All values denote success rate (\%).}
\label{tab:libero_metaworld}
\setlength{\tabcolsep}{5pt}
\begin{tabular}{llccccc |llccccc}
\hline
\multicolumn{7}{c}{LIBERO} & \multicolumn{7}{|c}{Meta-World} \\
\cline{1-7} \cline{8-14}
Method & Venue & Spatial & Object & Goal & Long & Average &
Method & Venue & Easy & Medium & Hard & Very Hard & Average \\
\hline
Octo~\cite{octo_2023} & RSS'24 & 78.9 & 85.7 & 84.6 & 51.1 & 75.1 &
DP3~\cite{Ze2024DP3}  & RSS'24 & 90.9 & 61.6 & 31.7 & 49.0 & 74.4 \\
QueST~\cite{mete2024quest} & NeurIPS'24 & \underline{89.0} & \underline{90.0} & \underline{88.4} & \underline{87.0} & \underline{88.6} &
Diffusion Policy~\cite{chi2025diffusion}  & IJRR'25 & 83.6 & 31.1 & 9.0 & 26.6 & 37.6 \\

MaIL~\cite{jia2024mail} & CoRL'24 & 53.8 & 81.5 & 56.3 & 41.7 & 58.3 &
Mamba Policy~\cite{cao2025mamba} & IROS'25 & \underline{95.4} & \underline{92.2} & \underline{48.3} & \underline{71.2} & \underline{76.8} \\
TraceVLA~\cite{zheng2025tracevla} & ICLR'25 & 84.6 & 85.2 & 75.1 & 54.1 & 74.8 &
TinyVLA-H~\cite{wen2024tinyvla} & ICRA'25 & 77.6 & 21.5 & 11.4 & 15.8 & 31.6 \\
\hline
\textbf{SUREFlow (Ours)} & IROS'26 & \textbf{94.8} & \textbf{91.0} & \textbf{93.8} & \textbf{90.2} & \textbf{92.5} & \textbf{SUREFlow (Ours)} & IROS'26 & \textbf{97.8} & \textbf{93.5} & \textbf{86.1} & \textbf{75.9} & \textbf{88.32} \\ \hline

\end{tabular}
\end{table*}

\section{Experimental Results}
\subsection{Experimental Setup}

\subsubsection{Implementation Details}
All models are implemented in \texttt{PyTorch} and trained with AdamW using a learning rate of $1\times10^{-4}$, momentum $0.9$, and weight decay $0.05$. Training is conducted for $300$ epochs with a batch size of $256$ on NVIDIA RTX 4090 GPU. For flow matching, the time variable $t \in [0,1]$ is uniformly sampled per batch, with noise $z \sim \mathcal{N}(0,I)$ used to construct $x_t$ as in Eq.~(2). During inference, we use $70$ flow integration steps. The time variable is provided as \texttt{sigma} and embedded via \texttt{TimeEmbedding}, corresponding to $\phi(t)$. The MGAD module uses 10 learnable action queries with a single-layer 4-head self and cross-attention decoder (head dim 64), followed by a 2-layer FFN (hidden 512) with LayerNorm. The output is linearly projected from 256 to 7 action dimensions. For perception, we use ResNet-18 as visual encoders (one per RGB view) with 256-dimensional latent projections, and a CLIP ViT-B/32 text encoder for language conditioning. Actions are encoded from $\mathbb{R}^{7}$ to the shared 
256-dimensional latent space using a linear embedding layer.
The Mamba backbone consists of $5$ layers with hidden dimension $256$ and intermediate dimension $256$. All token embeddings share a common embedding dimension of $256$. URFlow is enabled with $\lambda_u = 0.001$ without overpowering the primary flow-matching objective. During inference, refinement is triggered only when the predicted uncertainty exceeds $\tau = 0.9$, and each selected action is refined for 3 iterations. For ablation studies, Success Rate (SR) is averaged across all evaluation tasks, where each task is executed for $50$ rollouts.

\subsubsection{Dataset \& Models}
We evaluate SUREFlow on Meta-World~\cite{yu2020meta}, LIBERO~\cite{liu2023libero}, and LIBERO-PRO~\cite{zhou2025libero} to assess multi-task generalization, compositional reasoning, and long-horizon robustness. 
We compare SUREFlow with representative VLA and generative visuomotor policies spanning transformer, diffusion, and state-space architectures, including large-scale VLAs such as OpenVLA~\cite{kim2024openvla}, Octo~\cite{octo_2023}, and $\pi_{0.5}$~\cite{black2025pi}. All models follow their official training and evaluation protocols for fair comparison. Overall, our experiments aim to address the following key Research Questions (RQs):

\begin{itemize}
\item RQ1: Can URFlow reduce compounding integration errors in generative flow-based control?
\item RQ2: Does URFlow matching improve stability and success under extended rollout horizons?
\item RQ3: Can URFlow achieve long-horizon robustness comparable to large-scale VLA models, without relying on billion-parameter backbones?
\item RQ4: How does uncertainty calibration interact with selective refinement?
\end{itemize}

\begin{table*}[t]
\centering \scriptsize 
\caption{LIBERO-PRO model leaderboard showing normalized success rates under five perturbation types across four benchmarks.}
\label{tab:libero_pro_leaderboard}
\setlength{\tabcolsep}{2.8pt}
\renewcommand{\arraystretch}{0.95}
\begin{tabular}{lccccc |c ccccc |c ccccc |c ccccc |c|c}
\hline
\multirow{2}{*}{Model} & \multicolumn{5}{c|}{LIBERO Goal} & &
  \multicolumn{5}{c|}{LIBERO Spatial} & &
  \multicolumn{5}{c|}{LIBERO 10} & &
  \multicolumn{5}{c|}{LIBERO Object} & Average & \multirow{2}{*}{Params $\downarrow$}\\
  
\cline{2-6} \cline{8-12} \cline{14-18} \cline{20-24}
& Obj & Pos & Sem & Task & Env & &
  Obj & Pos & Sem & Task & Env & &
  Obj & Pos & Sem & Task & Env & &
  Obj & Pos & Sem & Task & Env & SR $\uparrow$ & \\
\hline
OpenVLA~\cite{kim2024openvla}
& 0.96 & 0.00 & 0.98 & 0.00 & 0.98 & &
  0.97 & 0.00 & 0.97 & 0.00 & {0.89} & &
  0.81 & 0.00 & 0.96 & 0.00 & 0.85 & &
  0.98 & 0.00 & {0.98}& 0.00 & 0.00 & \underline{0.52} & $7$B\\

$\pi_{0}$~\cite{black2024pi_0}
& 0.94 & 0.00 & 0.93 & 0.00 & 0.39 & &
  0.95 & 0.00 & 0.97 & 0.00 & 0.60 & &
  0.79 & 0.00 & 0.82 & 0.00 & 0.27 & &
  {0.94} & 0.00 & 0.90 & 0.00 & 0.29 & 0.44 & $3$B\\

$\pi_{0.5}$~\cite{black2025pi}
& 0.97 & 0.38 & 0.97 & 0.00 & 0.46 & &
  0.97 & 0.20 & 0.97 & 0.01 & 0.46 & &
  0.92 & 0.08 & 0.93 & 0.01 & 0.46 & &
  {0.98} & {0.17} & {0.96} & 0.01 & {0.73}& \textbf{0.53} & $3$B\\
\hline
    \textbf{SUREFlow (Ours)} & 
    0.93 & 0.00 & 0.89 & 0.00 & 0.93 & &
    0.92 & 0.00 & 0.90 & 0.00 & 0.93 & &
    0.21 & 0.00 & 0.78 & 0.00 & 0.74 & &
    0.68 & 0.00 & 0.94 & 0.00 & 0.91 & 0.49 & \textbf{179.1M}\\ \hline
\end{tabular}
\end{table*}

\subsection{Comparison with SOTA Methods}

\begin{figure}[!t]
    \centering
    \includegraphics[width=0.48\textwidth]{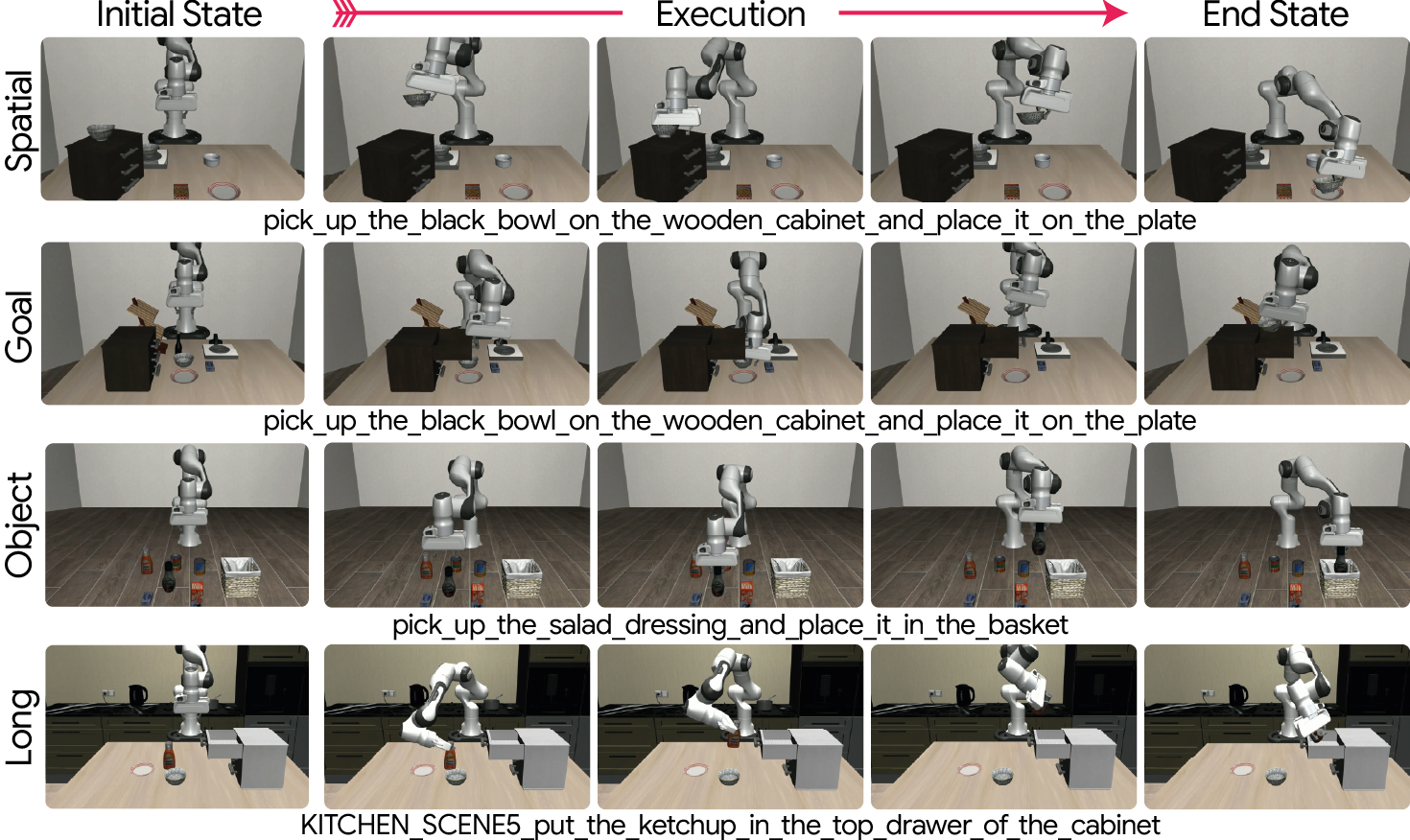}
   \caption{Example evaluation outcome of our SUREFlow on LIBERO.}
    \label{fig:simulation}
\end{figure}

To address RQ1 and RQ2, we first evaluate SUREFlow on the LIBERO and Meta-World benchmarks. 
As summarized in Table I, our SUREFlow achieves an average SR of 92.50\%, outperforming transformer-based models like Octo by 17.4\% and SSM-based methods MaIL by 34.2\%. Representative simulation results across the LIBERO suites are presented in Fig.~\ref{fig:simulation}. On the Meta-World, which emphasizes diverse manipulation skills with varying difficulty levels, SUREFlow achieves an average SR of 88.32\%, surpassing DP3 at 74.4\%, Diffusion Policy at 37.6\%, Mamba Policy at 76.8\%, and TinyVLA-H at 31.6\%. Notably, compared to Diffusion Policy, SUREFlow improves average success by nearly 78\%, highlighting the capability of our SUREFlow's efficient and stable action generation.

Moreover, 34.2\% gain over MaIL and 11.52\% improvement over Mamba Policy, which are both based on Mamba backbone, further confirms that the performance gains are not solely attributable to the State-space backbone, but arise from our proposed URFlow-based uncertainty-aware residual modeling and closed-loop refinement. Thus, integrating URFlow within a state-space backbone improves long-horizon stability and robustness, directly addressing RQ1.

 Unlike autoregressive transformer VLAs that depend on large-scale pretrained backbones, SUREFlow learns a conditional velocity field with explicit uncertainty calibration, enabling selective refinement of unreliable action dimensions during rollout. The strong performance across easy to very hard task categories, together with improvements under extended horizons, directly supports RQ2 by demonstrating that uncertainty-guided refinement enhances stability under increasing task complexity and rollout length.

\subsection{Robustness Under Distribution Shift and Efficiency}

To evaluate robustness under distribution shift and address RQ3, we compare SUREFlow with large-scale VLAs on LIBERO-PRO, which measures normalized SR under object position, semantic, task, and environment perturbations. Our SUREFlow achieves an average SR of 0.49 using $179.1$M parameters. While OpenVLA ($7$B) and $\pi_{0.5}$ ($3$B) obtain 0.52 and 0.53, respectively, SUREFlow remains competitive despite being more than an order of magnitude smaller. It also outperforms $\pi_{0}$ ($3$B), which achieves an SR of 0.44.

Despite operating with only $179.1$M parameters, SUREFlow attains performance close to OpenVLA ($7$B) and $\pi_{0.5}$ ($3$B) under diverse perturbation settings, indicating that URFlow effectively compensates for model scale. 
The Mamba backbone in our SUREFlow enables linear-time modeling without quadratic transformer cost, while uncertainty-guided refinement stabilizes extended-horizon rollouts and reduces compounding errors. These results directly address RQ3 by demonstrating that our model SUREFlow achieves long-horizon robustness comparable to large VLA models.

\begin{figure*}[t]
    \centering
    \includegraphics[width=1.0\textwidth]{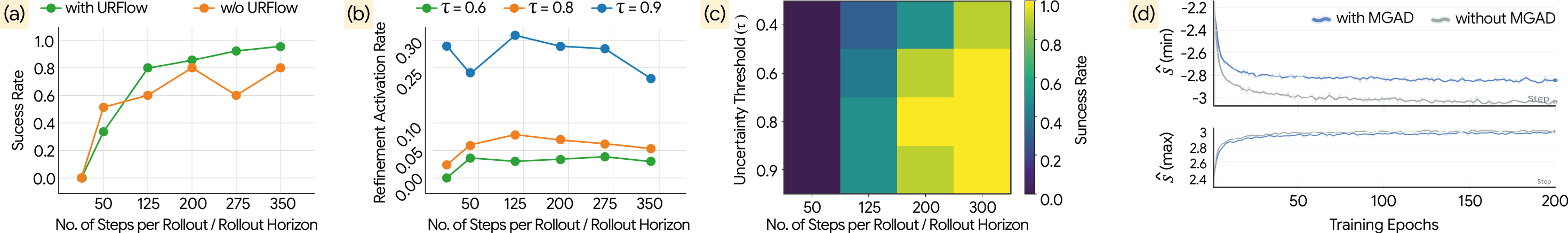}
   \caption{Long-horizon rollout stability and URFlow-based refinement analysis on LIBERO Long. (a) Success rate vs. rollout horizon comparing SUREFlow with and without URFlow, showing slower performance degradation under extended rollouts when uncertainty-guided refinement is enabled.
(b) Refinement activation rate under different uncertainty thresholds $\tau \in \{0.6,0.8,0.9\}$ as rollout horizon increases; lower $\tau$ triggers more frequent updates, while higher $\tau$ yields more selective refinement.
(c) Heatmap of success across thresholds and horizons, illustrating stable performance for moderate $\tau$ under long-horizon settings.
(d) Evolution of predicted uncertainty statistics during training, indicating the stability with and without our MGAD module.}
    \label{fig:questions}
\end{figure*}

\subsubsection{Stability and Uncertainty Calibration Analysis}
 To further address RQ2, RQ3, and RQ4, we analyze long-horizon rollout behavior and uncertainty dynamics in Fig.~4. An extended rollout is particularly prone to compounding errors during open-loop execution, where small velocity inaccuracies accumulate over time. 
 Fig. 4(a) shows that with URFlow enabled, performance improves steadily and remains stable as the rollout horizon increases.
In contrast, without URFlow, the success rate decreases more rapidly and becomes less stable as the number of steps grows.
 This directly supports RQ2 by demonstrating that URFlow matching improves stability and success under extended rollout horizons. Furthermore, the improved stability under increasing horizons achieved without scaling model size further supports RQ3. This is because URFlow achieves long-horizon rollout robustness through calibrated residual refinement rather than reliance on billion-parameter backbones. 
Fig.~4(b) shows refinement activation rates under variable uncertainty thresholds $\tau$. Lower $\tau$ values trigger more frequent residual updates, whereas higher values yield more selective refinement. The corresponding success heatmap in Fig.~4(c) reveals that moderate thresholds maintain stable performance across longer horizons, while overly aggressive or overly conservative refinement leads to reduced gains. These results address RQ4 by showing that uncertainty calibration directly governs the trade-off between refinement frequency and stability, enabling selective correction of unreliable action dimensions without introducing excessive updates.

\subsubsection{Effect of MGAD on Uncertainty Calibration}
Fig.~4(d) shows the evolution of predicted uncertainty statistics during training, comparing models with and without MGAD. The trends of $\hat{s}_{\min}$ and $\hat{s}_{\max}$ indicate that MGAD leads to faster convergence and a more stable uncertainty range. In contrast, removing MGAD results in consistently lower $\hat{s}_{\min}$ values, suggesting overconfident predictions. By moderating extreme values and maintaining a bounded uncertainty distribution, MGAD improves calibration and contributes to more stable behavior in extended rollout.

\begin{table}[t]
\centering \scriptsize 
\caption{Ablation study on SUREFlow components.}
\label{tab:ablation}
\setlength{\tabcolsep}{1.7pt}
\renewcommand{\arraystretch}{0.95}
\begin{tabular}{lccc ccc}
\hline
Variant & FCM & MGAD & URFlow & SR$\uparrow$ & Params (M)$\downarrow$ & FLOPs (G)$\downarrow$ \\
\hline
+ FCM & $\checkmark$ & $\times$ & $\times$ & 0.72 & 178.30 & 6.48 \\
+ MGAD & $\times$ & $\checkmark$ & $\times$ & 0.77 & 178.57 & 6.78 \\
+ URFlow & $\times$ & $\times$ & $\checkmark$ & 0.81 & 177.78 & 6.41 \\ 
+ FCM + MGAD & $\checkmark$ & $\checkmark$ & $\times$ & 0.85 & 179.10 & 6.85 \\
+ MGAD + URFlow & $\times$ & $\checkmark$ & $\checkmark$ & 0.91 & 178.57 & 6.78 \\
+ FCM + URFlow & $\checkmark$ & $\times$ & $\checkmark$ & 0.87 & 178.30 & 6.48 \\ \midrule
\textbf{SUREFlow (Baseline)} & \textbf{$\checkmark$} & \textbf{$\checkmark$} & \textbf{$\checkmark$} & 0.92 & 179.10 & 6.85 \\
\hline
\end{tabular}
\end{table}

\begin{table}[t]
\centering \scriptsize
\caption{Sensitivity analysis of URFlow on LIBERO (average SR).}
\vspace{-0.2cm}
\label{tab:URFlow_ablation}
\setlength{\tabcolsep}{6pt}
\begin{tabular}{l|lc|l|lc}
\hline
Effect of &Configuration & SR$\uparrow$ &  Effect of & Configuration & SR$\uparrow$\\ \hline

Uncertainty & $\lambda_u = 0.001$ & \textbf{0.92} & {Refinement} & $K = 2$ & 0.85  \\
{weight} &$\lambda_u = 0.01$ & 0.87 & {steps} & $K = 3$ & \textbf{0.92} \\
$\lambda_u$&$\lambda_u = 0.05$ & 0.79 & $K$ & $K = 5$ & 0.88 \\
\hline
\end{tabular}
\end{table}

\subsection{Ablation Study}

\subsubsection{Component Analysis}
Table III evaluates the impact of FCM, MGAD, and URFlow. Individually, FCM achieves 0.72 SR, MGAD 0.77, and URFlow 0.81, with URFlow providing the greatest single-module improvement. Combining FCM and MGAD increases SR to 0.85, while MGAD with URFlow reaches 0.91, indicating strong complementary effects. The full SUREFlow model achieves the highest SR of 0.92. Notably, these improvements are obtained with only modest changes in parameter count and FLOPs, confirming that performance gains arise from improved modeling rather than brute-force scaling.

\subsubsection{Sensitivity Analysis of URFlow}
It is observed from the results in Table~IV that moderate weighting improves stability and success. The best result of 0.92 is achieved with a properly calibrated $\lambda_u$, whereas excessive weighting degrades performance, emphasizing the need for balanced uncertainty regularization. Similarly, refinement depth influences robustness: 2 refinement steps achieve 0.85 with moderate variance, while 3 steps attain 0.92 with improved stability; additional steps provide no further gains. These results confirm that properly calibrated uncertainty weighting and selective refinement are critical for mitigating compounding errors without overcorrection.

\begin{figure}[t]
    \centering
    \includegraphics[width=0.48\textwidth]{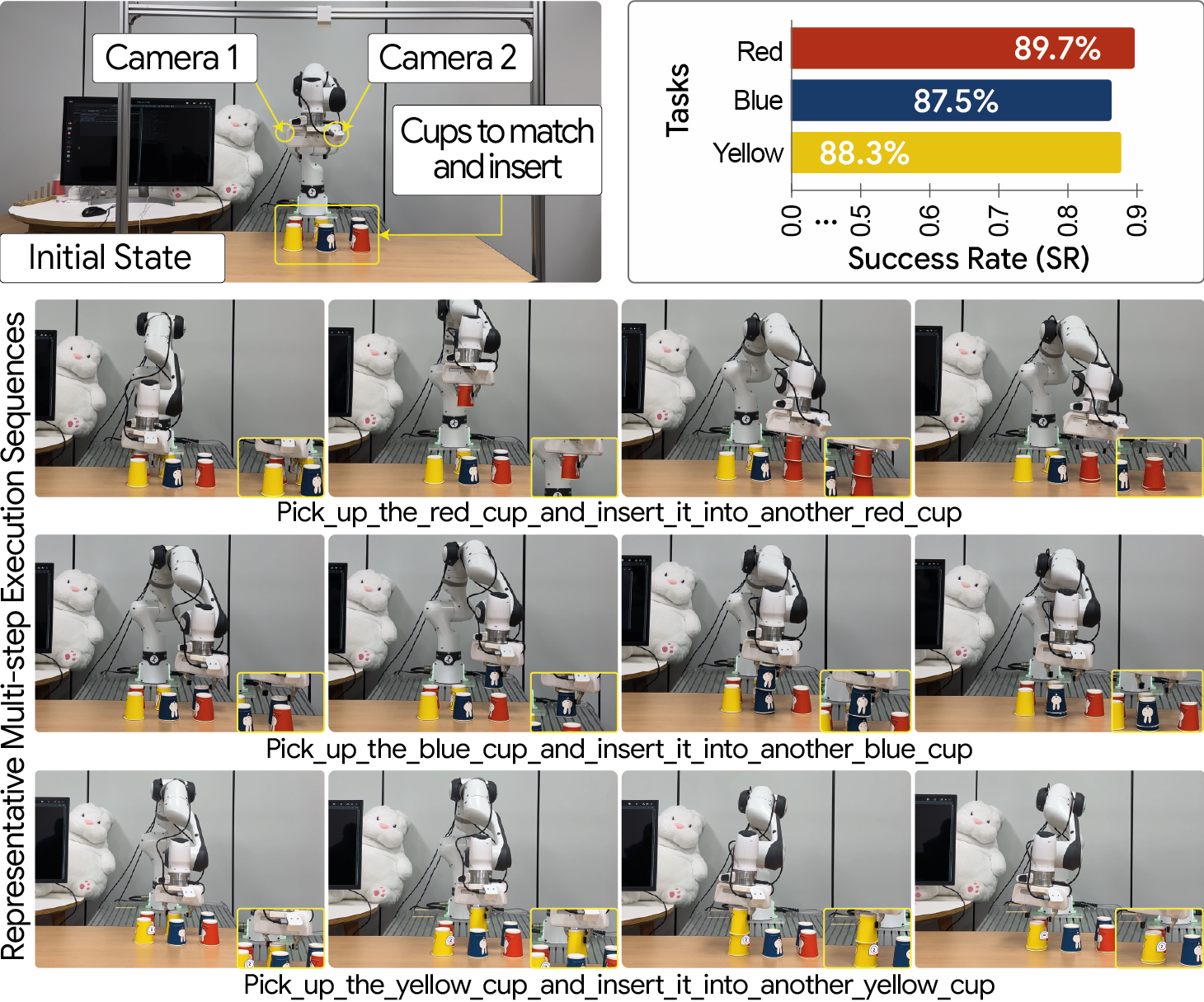}
    \caption{Real-robot color-matching \& cup insertion task by SUREFlow. Top: two-camera setup and SR over 20 rollouts per color. Bottom: representative multi-step execution sequences for red, blue, and yellow cups.}
    \label{fig:real_robot}
\end{figure}
\section{Real Robot Experiment}
We evaluate SUREFlow on a real-world tabletop manipulation task using a Franka Emika arm with a parallel gripper and an Intel RealSense D405 RGB-D camera, with inference running on an NVIDIA RTX 4090 GPU. The setup includes $6$ cups in $3$ colors, red, yellow, and blue, with two cups per color. Given a visual observation, the robot is required to (1) select one cup, (2) identify the second cup of the same color, and (3) insert the grasped cup onto its color-matched counterpart. This setup requires reliable color discrimination, precise grasping, and accurate spatial alignment during insertion. A total of 20 independent rollouts were conducted across the three color pairs. 

SUREFlow achieved an average SR of 88.5\% over 20 rollouts per color (Fig.~\ref{fig:real_robot}), consistently completing the color-matching pick-and-insert task. The model demonstrates reliable color-conditioned reasoning, stable grasping, and accurate insertion alignment. These results suggest that uncertainty-aware residual refinement enhances robustness under real-world perception noise and minor pose variations, supporting the practical effectiveness of SUREFlow.

\section{Conclusion}
We presented SUREFlow, a state-space uncertainty-aware residual flow matching framework for robust generative robot manipulation. By integrating URFlow with a Mamba-based backbone and MGAD, our method models input-dependent uncertainty and performs selective residual refinement to mitigate compounding errors during extended rollouts. Extensive evaluations on LIBERO, Meta-World, and LIBERO-PRO demonstrate strong performance gains over both transformer-based and state-space baselines, while maintaining a lightweight design. Real-world experiments further validate the practical robustness of our SUREFlow.

\section*{Acknowledgments}
This work was supported by the National Research Foundation of Korea (NRF) grant funded by the Korea government(MSIT)(No. RS-2025-02214941). This research was supported by the Regional Innovation System \& Education(RISE) Glocal 30 program through the Daegu RISE Center, funded by the Ministry of Education(MOE) and the Daegu, Republic of Korea (2025-RISE-03-001).










\bibliographystyle{ieeetr}
\bibliography{03_ref}

\end{document}